\documentclass[runningheads]{llncs}
\usepackage{graphicx}

\usepackage{tikz}
\usepackage{comment}
\usepackage{amsmath,amssymb} 
\usepackage{color}
\usepackage{soul}

\begin{document}
\pagestyle{headings}
\mainmatter
\def\ECCVSubNumber{23}  

\title{On-the-go Reflectance Transformation Imaging with Ordinary Smartphones} 


\titlerunning{On-the-go RTI with Ordinary Smartphones}
%
\author{Mara Pistellato\inst{1}\orcidID{0000-0001-6273-290X} \\ \and 
Filippo Bergamasco\inst{1}\orcidID{0000-0001-6668-1556}}
\authorrunning{M. Pistellato et al.}
%
\institute{DAIS, Università Ca'Foscari Venezia\\ 155, via Torino, Venezia Italy \\
\email{\{mara.pistellato,filippo.bergamasco\}@unive.it}}
\maketitle

\begin{abstract}
Reflectance Transformation Imaging (RTI) is a popular technique that allows the recovery of per-pixel reflectance information by capturing an object under different light conditions.
This can be later used to reveal surface details and interactively relight the subject.
Such process, however, typically requires dedicated hardware setups to recover the light direction from multiple locations, making the process tedious when performed outside the lab.
\\
We propose a novel RTI method that can be carried out by recording videos with two ordinary smartphones. The flash led-light of one device is used to illuminate the subject while the other captures the reflectance. Since the led is mounted close to the camera lenses, we can infer the light direction for thousands of images by freely moving the illuminating device while observing a fiducial marker surrounding the subject.
To deal with such amount of data, we propose a neural relighting model that reconstructs object appearance for arbitrary light directions from extremely compact reflectance distribution data compressed via Principal Components Analysis (PCA).
Experiments shows that the proposed technique can be easily performed on the field with a resulting RTI model that can outperform state-of-the-art approaches involving dedicated hardware setups. 
\end{abstract}

\section{Introduction}

In Reflectance Transformation Imaging (RTI) an object is acquired with different known light conditions to approximate the per-pixel Bi-directional Reflectance Distribution Function (BRDF) from a static viewpoint.
Such process is commonly used to produce relightable images for Cultural Heritage applications \cite{mytum2018application,earl2011reflectance} or perform material quality analysis \cite{coules2019reflectance} and surface normal reconstruction.
The flexibility of such method makes it suitable for several materials, and the resulting images can unravel novel information about the object under study such as manufacturing techniques, surface conditions or conservation treatments.
Among the variety of practical applications in Cultural Heritage field, we can mention enhanced visualisation \cite{earl2011reflectance,palma2010dynamic}, documentation and preservation \cite{manrique2013applications,kotoula2013study,manfredi2013measuring} as well as surface analysis \cite{ciortan2016practical}.
Moreover, RTI techniques can be effectively paired with other tools as 3D reconstruction \cite{uribe2013rock,Pistellato20163703,Pistellato2015468,Pistellato2019640} or multispectral imaging \cite{giachetti2017multispectral} to further improve the results.

In the majority of the cases, the acquisition of RTI data is carried out with specialised hardware involving a light dome and other custom devices that need complex initial calibration.
Since the amount of processed data is significant, several compression methods have been proposed for RTI data representation to obtain efficient storage and interactive rendering \cite{ponchio2019relight,giachetti2018novel}.
In addition to that, part of the proposals focus on the need of low-cost portable solutions \cite{kinsman2016easy,watteeuw2016light,porter2016portable}, including mobile devices \cite{schuster2014rti} to perform the computation on the field.

In this paper we first propose a low-cost acquisition pipeline that requires a couple of ordinary smartphones and a simple marker printed on a flat surface.
During the process, both smartphones acquire two videos simultaneously: one device acting as a static camera observing the object from a fixed viewpoint, while the other provides a trackable moving light source.
The two videos are synchronised and then the marker is used to recover the light position with respect to a common reference frame, originating a sequence of intensity images paired with light directions.
The second contribution of our work is an efficient and accurate neural-network model to describe per-pixel reflectance  based on PCA-compressed intensity data.
We tested the proposed relighting approach both on a synthetic RTI dataset, involving different surfaces and materials, and on several real-world objects acquired on the field.

\section{Related Work}

The literature counts a huge number of different methods for both acquisition and processing of RTI data for relighting.
In \cite{pintus2019state} the authors give a comprehensive survey on Multi-Light Image Collections (MLICs) for surface analysis.
Many approaches employ the classical polynomial texture maps \cite{malzbender2001polynomial} to (i) define the per-pixel light function, (ii) store a representation of the acquire data, and (iii) dynamically render the image under new lights.
Similar techniques are the so-called Hemispherical Harmonics coefficients \cite{mudge2008image} and Discrete Modal Decomposition \cite{pitard2017discrete}.
In \cite{giachetti2018novel} the authors propose a new method based on Radial Basis Function (RBF) interpolation, while in \cite{ponchio2019relight} a compact representation for web visualisation employing PCA is presented.
The authors in \cite{mudge2006new} present the Highlight Reflectance Transformation Imaging (H-RTI) framework, where the light direction is estimated by detecting its specular reflection on one or more spherical objects captured in the scene.
However, such setup involves several assumptions such as constant light intensity and orthographic camera model, that in practice make the model unstable.
Other techniques that have been proposed to estimate light directly from some scene features are \cite{ackermann2013geometric,ahmad2014improved}, while in the authors \cite{giachetti2018novel} propose a novel framework to expand the H-RTI technique.

Recently, neural networks have been employed successfully in several Computer Vision tasks, including RTI.
In particular, the encoder-decoder architecture is used in several applications for effective data compression \cite{smys2020survey}.
The work in \cite{ren2015image} presents a NN-based method to model light transport as a non-linear function of light position and pixel coordinates to perform image relighting.
Other related work using neural networks are \cite{xu2018deep}, in which a subset of optimal light directions is selected, and \cite{rainer2019neural} where a convolutional approach is adopted.
Authors in \cite{dulecha2020neural} propose an autoencoder architecture to perform relighting of RTI data: the architecture is composed by an encoder part where pixel-wise acquired values are compressed, then the decoder part uses the light information to output the expected pixel value. They also propose two benchmark datasets for evaluation.


\section{Proposed Method}

\begin{figure}[t]
    \centering
    \includegraphics[width=\linewidth]{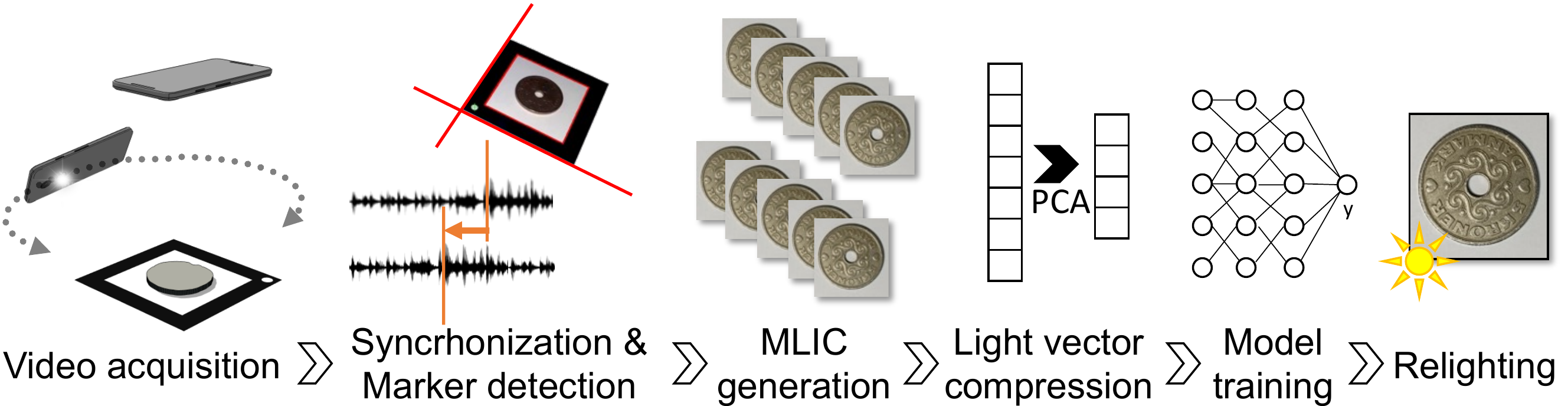}
    \caption{Complete mobile-based RTI acquisition and relighting pipeline.}
    \label{fig:pipeline}
\end{figure}

Our method follows the classical procedure employed in the vast majority of existing RTI applications: the whole pipeline is presented in Figure \ref{fig:pipeline}.
First, several images of the object under study are acquired varying the lighting conditions. In our case, the operation uses the on-board cameras and flash light of a pair of ordinary smartphones while taking two videos.
The two videos are then synchronised and the smartphones positions with respect to the scene are recovered using a fiducial marker: in this way we obtain light position and reflectance image for each frame.  Such data is processed to create a model that maps each pair (\emph{pixel}, \emph{light direction}) to an observed reflectance value. Section \ref{sec:data_acquisition} gives a detailed description of this process.
This results in a Multi-Light Image Collection (MLIC), that is efficiently compressed by projecting light vectors to a lower-dimensional space via PCA. Then, we designed a neural model defined as a small Multi-Layer Perceptron (MLP) to decode the compressed light vectors and extrapolate the expected intensity of a pixel given a light direction.
In Section \ref{sec:reflectance_model} the neural reflectance model and data compression are illustrated in detail.
Finally, the trained model is used to dynamically relight the object by setting the light direction to any (possibly unseen) value.

\subsection{Data Acquisition}\label{sec:data_acquisition}

\begin{figure}
    \centering
    \includegraphics[height=6cm]{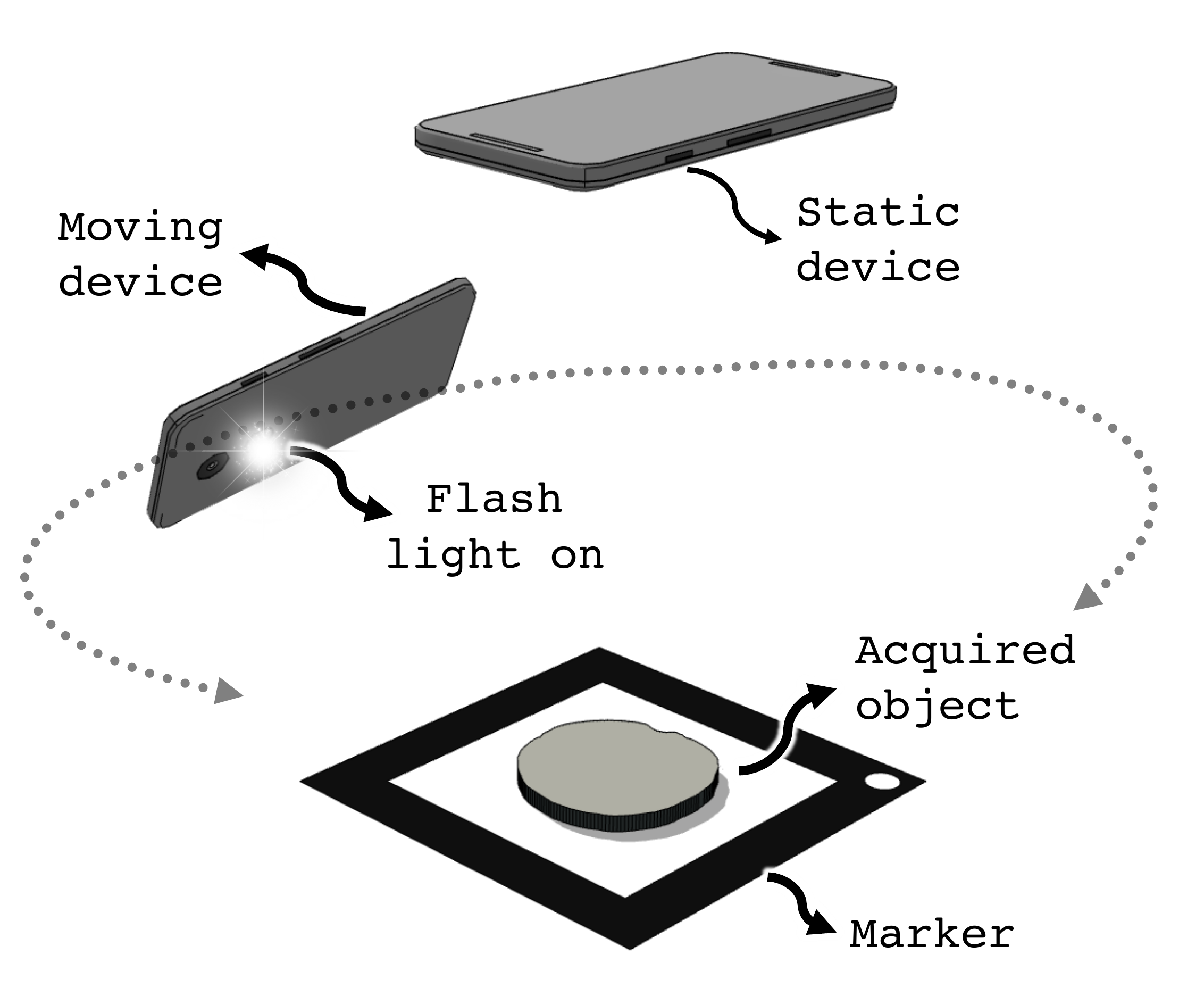}
    \includegraphics[height=6cm]{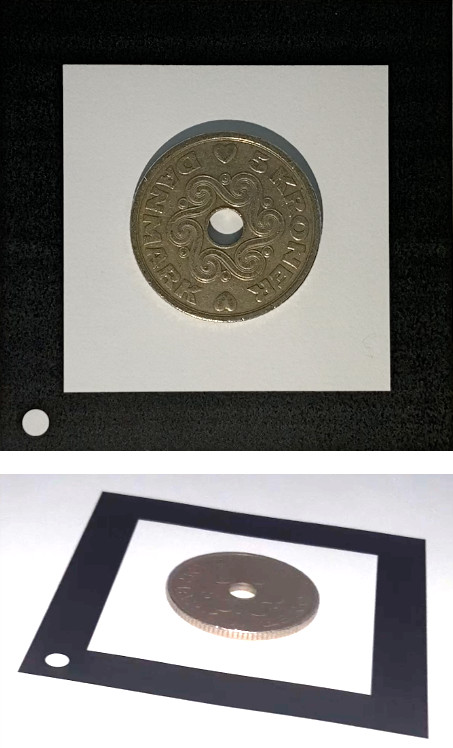}
    \caption{Left: Proposed RTI acquisition setup. Right: Example frames acquired by the static and moving devices.}
    \label{fig:setup}
\end{figure}

Data acquisition is performed using two smartphones and a custom fiducial marker as shown in Figure \ref{fig:setup} (left). The object to acquire is placed at the centre of a marker composed by a thick black square with a white dot at one corner. 

One device is located above the object, with the camera facing it frontally so that it produces images as depicted in Figure \ref{fig:setup} (top-right). This device, called \emph{static}, must not move throughout the acquisition, so we suggest to attach it to a tripod.
The second device, called \emph{moving}, is manually moved around the object with an orbiting trajectory. The flash led-light located close to the backward-facing camera must be kept on all the time to illuminate the object from different locations. This will allow the static device to observe how the reflectance of each pixel changes while moving the light source.

Both the devices record a video during the acquisition. For now, let's consider those videos as just sequences of images perfectly synchronised in time. In other words, the acquisition consists in a sequence of $M$ images $(I_0^s, I_1^s, \ldots, I_M^s)$ acquired from the static device paired with a sequence $(I_0^m, I_1^m, \ldots, I_M^m)$ acquired from the moving device at times $t_0, t_1, \ldots, t_M$.

After video acquisition, each image is processed to detect the fiducial marker. For the static camera, this operation is needed to locate the $4$ corners $(c_0, c_1, c_2, c_3)$ of the inner white square (i.e. the internal part of the marker inside the thick black border). This region is then cropped to create a sequence of $(\mathcal{I}_0, \ldots, \mathcal{I}_N)$ images composed by $W {\times} H$ pixels commonly referred as Multi-light Image Collection (MLIC). Note that $N$ can be lower than $M$ because the fiducial marker must be detected in both $I_i^s$ and $I_i^m$ to be added to the MLIC. 

Each $\mathcal{I}_i$ is a single-channel grayscale image containing only the luminance of the original $I_i^s$ image. We decided to model only the reflectance intensity (and not the wavelength) as a function of the light's angle of incidence for two reasons. First, we cannot change the colour of the light source and, second, it is uncommon to have iridescent materials where the incindent angle affects the reflectance spectrum~\cite{Kinoshita_2008}. Therefore, we convert all the images to the YUV colour space to store only the Y channel in the MLIC. To deal with the colour, we store the pixel-wise averages $\bar{U} = \frac{1}{N}\sum_N U_i$ and $\bar{V} = \frac{1}{N}\sum_N V_i$ for further processing.

The marker is also detected in the moving camera image sequence, but for a different purpose. We assume that the flash light is so close to the camera optical centre that can be considered almost at the same point. So, by finding the pose of the camera $(R, t)$ in the marker reference frame, we can estimate the location of the light source (i.e. the moving camera optical center) with respect to the object. This operation is simply performed by computing the Homography $H$ mapping $c_0 \ldots c_3$ to the marker model points $\begin{pmatrix}0\\0\\1 \end{pmatrix},\begin{pmatrix}W\\0\\1 \end{pmatrix},\begin{pmatrix}W\\H\\1 \end{pmatrix},\begin{pmatrix}0\\H\\1 \end{pmatrix}$ and then factorising it as:

\begin{equation}
    K^{-1} H = \alpha \begin{pmatrix} \vert & \vert & \vert \\ r_1 & r_2 & t \\ \vert & \vert & \vert \end{pmatrix}
\end{equation}

\noindent where $K$ is the intrinsic camera matrix, $r_1, r_2$ are the first two columns of the rotation matrix $R$, and $\alpha$ is a non-zero unknown scale factor~\cite{Hartley2003}.
Since $R$ must be orthonormal, $\alpha$ can be approximated as $2/( \| r_1 \| + \| r_2 \| )$ and $r_3$ as $r_1 \times r_2$. Since $(R,t)$ maps points from the camera reference frame to the marker (i.e. object) reference frame, the vector $t$ represents the light position and $\mathcal{L}=t/\| t \|$ the light direction. Since the light is not at the infinity, each object point actually observes a slightly different light direction vector $\mathcal{L}$.
However, as usually done in other RTI applications, we consider this difference negligible so that we collect a single light direction vector for each image.

After the data acquisition process, we end up with the MLIC $(\mathcal{I}_0 \ldots \mathcal{I}_N)$ and $(\mathcal{L}_0, \ldots, \mathcal{L}_N)$ vectors together with $\bar{U}, \bar{V}$. This is all the data we need to generate our reflectance model and proceed with dynamic relighting.
At this point, we need to do some considerations regarding the acquisition procedure: 
\begin{itemize}
    \item Pixel reflectance data is collected only from the static camera, while the moving camera is used just to estimate the light direction. This implies that the final result quality is directly affected by the quality of the static camera (i.e. resolution, noise, etc.). Therefore, we suggest to use a good smartphone for that. Conversely, the moving device can be cheap as long as images are sufficiently well exposed to reliably detect the marker.
    \item The moving camera must be calibrated a-priori to factorise H. In practice, the calibration is not critical and can simply be inferred from the lens information provided in the EXIF metadata. We also used this approach in all our experiments. 
    \item The ambient should be illuminated uniformly and constantly over time. Ideally, the moving device flash light should be the only one observed by the object. Since that is typically impractical, it is at least sufficient that the contribution from ambient illumination is negligible with respect to the provided moving light. 
    \item The orbital motion should uniformly span the top hemisphere above the object with a certain constant radius. Indeed, we only consider light direction so changes in the reflectance due to light proximity with respect to the object will not be properly accounted by the model.
\end{itemize}

\subsubsection{Video Synchronisation}

Video recording is manually started (roughly at the same time) in the two devices. So, it is clear that the two frame sequences are not synchronised out of the box (i.e. the $i^{th}$ frame of the static device will not be taken at the same time as the $i^{th}$ frame of the moving one). That will never be the case without an external electronic triggering but we can still obtain a reasonable synchronisation exploiting the audio signal of the two videos~\cite{vieira2020low}. 

We first extract the two audio tracks and then compute the time offset in seconds that maximises the Time-lagged Cross-correlation~\cite{SHEN2015680}. Once the offset is known, initial frames from the video starting first are dropped to match the two sequences. Note also that, if the framerates are different, frames must be dropped from time to time from the fastest video to keep it in sync with the other.
In the worst case, the time skewness is 1/FPS where FPS is the framerate of the slowest video. Nevertheless, since the moving device is orbited around the object very slowly, such time skewness will have a negligible effect in the estimation of $(\mathcal{L}_0, \ldots, \mathcal{L}_N)$.

\subsubsection{Fiducial Marker Detection}\label{sec:fiducial_detection}

Detecting the four internal corners of the proposed fiducial marker can be simply performed with classical image processing. We start with Otsu's image thresholding~\cite{otsu79} followed by hierarchical closed contour extraction~\cite{suzuki85}. Each contour is then simplified with the Ramer-Douglas-Peucker algorithm and filtered out if resulting in a number of points different than $4$. All the black-to-white 4-sides polygons contained into a white-to-black 4-side polygon are good candidates for a marker. So, we check the midpoint of each closest corresponding vertex pairs searching for the white dot. If exactly one white dot is found among corresponding pairs, than the four vertexes of the internal polygon can be arranged in clockwise order starting from the one closest to the dot. This results in the four corners $c_0, \ldots, c_3$.

Since we expect to see exactly one instance of the marker in every frame, this simple approach is sufficient in practice. We decided not use popular alternative markers (see for instance~\cite{aruco}) because they typically reserve the internal payload area to encode the marker id. Of course, any method will work as long as it results in a reasonably accurate localisation of the camera while providing free space to place the object under study.

\subsection{The Reflectance Model}\label{sec:reflectance_model}

To perform interactive relighting, we need first to model how the reflectance changes when varying the light direction. Our goal is to define a function: 

\begin{equation}
f\big( \mathbf{p}, \vec{l}\big) \to (y,u,v)
\end{equation}

\noindent producing the intensity $y$ and colour $u,v$ (in the YUV space) of a pixel $\mathbf{p}=(x,y) \in \mathbb{N} {\times} \mathbb{N}$ when illuminated from a light source with direction $\vec{l}=(l_u, l_v)$\footnote{$l_u$ and $l_v$ range between $[-1 \ldots 1]$ respectively as they are the first two components of a (unitary-norm) 3D light direction vector pointing toward the light source.}. Once the model is known, relighting can be done by choosing a light $\vec{l}$ and evaluating $f$ for every pixel $\mathbf{p}$ of the target image.

The data acquisition process described before produces a sampling of $f$ for some discrete values of $\mathbf{p}$ and $\vec{l}$. This sampling is dense on the pixels, since we acquire an entire image for every light, but typically sparse in in the amount of the observed light directions, especially if using a light dome where this number is limited to a few dozens. In our case, the sampling of $\vec{l}$ is a lot denser since we acquire an entire video composed by thousands of frames. However, directions are highly correlated in space as we follow a continuous circular trajectory (See Figure~\ref{fig:network_and_lights}, Left).

The challenge is to: (i) provide a realistic approximation of $f$ for previously unseen light directions while (ii) using a very compact representation so that it can be easily transferred, stored and evaluated even on a mobile phone. The two problems are related because the selection of what family of functions to use for $f$ affects how many parameters are needed to describe the chosen one. For example, in~\cite{malzbender2001polynomial} each pixel is independently modelled as a 6-coefficients biquadratic function of the light direction, requiring the storage of $6{\times}M{\times}N$ values.

Inspired by \emph{NeuralRTI} proposed by Dulecha et. al. \cite{dulecha2020neural}, we also represent $f$ as a Multi-Layer Perceptron trained from the data acquired with the smartphones. However, we have two substantial differences with respect to their approach.
First, we avoid the auto-encoder architecture for data compression. Since we do not use a light dome, the number of light samples changes in each acquisition and is at least an order of magnitude greater. NeuralRTI would not be feasible in our case, as it results in a network taking in input vectors of thousands of elements. Also, the network architecture itself depends on $N$ (variable in our case) producing a different layout for each acquired object. Instead, we compress such vectors with classical PCA to feed the MLP acting as a decoder. Interestingly, this tend to produce better results not only on our data but also on images acquired with a classical light dome.
Second, the light vector $\vec{l}$ is not concatenated as-is to the network input but projected to a higher-dimensional Fourier space with random frequencies as discussed in~\cite{tancik2020fourier}. This has a positive effect on the ability of the network to reconstruct the correct pixel intensity.

\subsection{Neural Model}

\begin{figure}
    \centering
    \includegraphics[height=4cm]{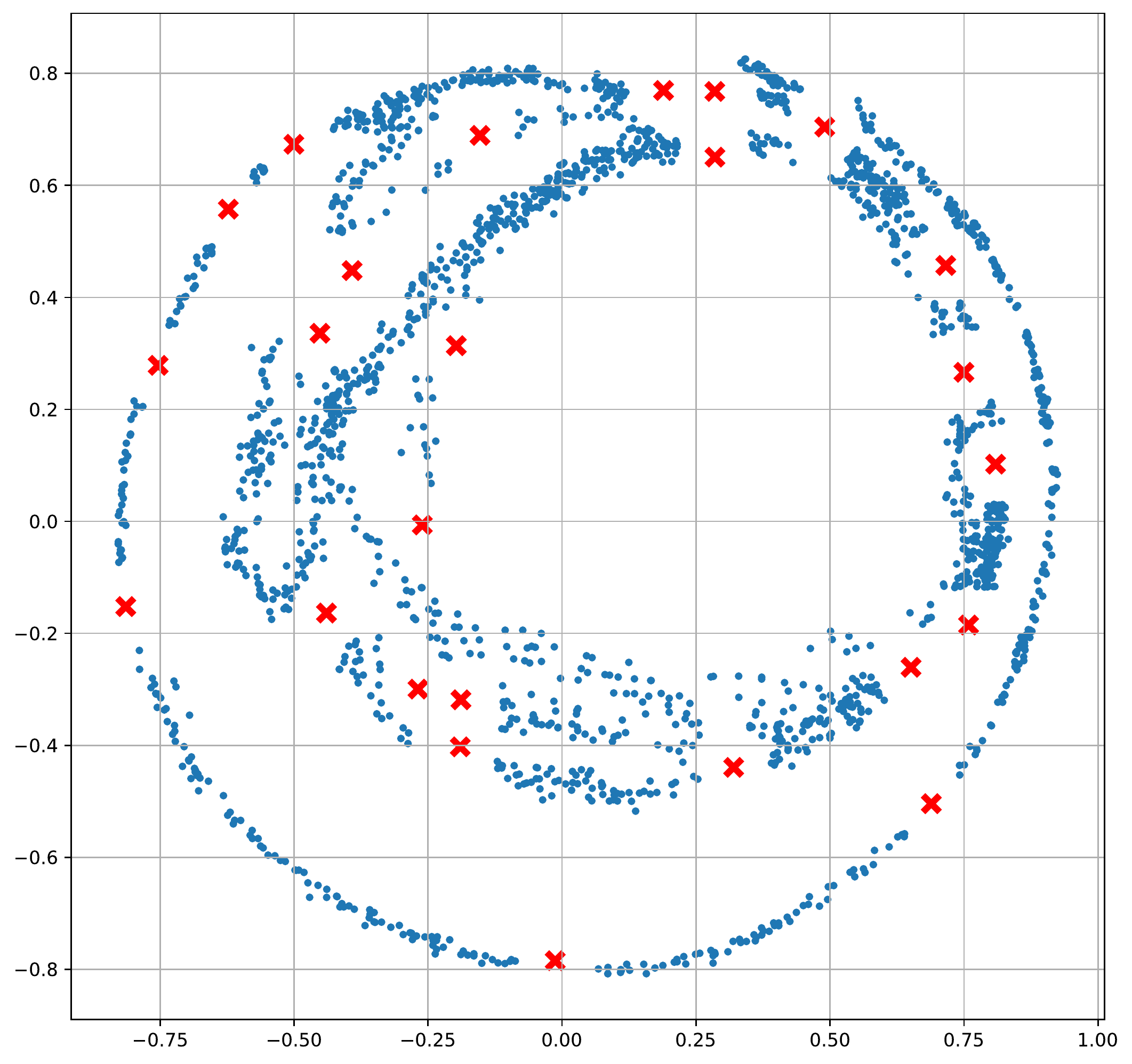}\hspace{0.1cm}
    \includegraphics[height=4.5cm]{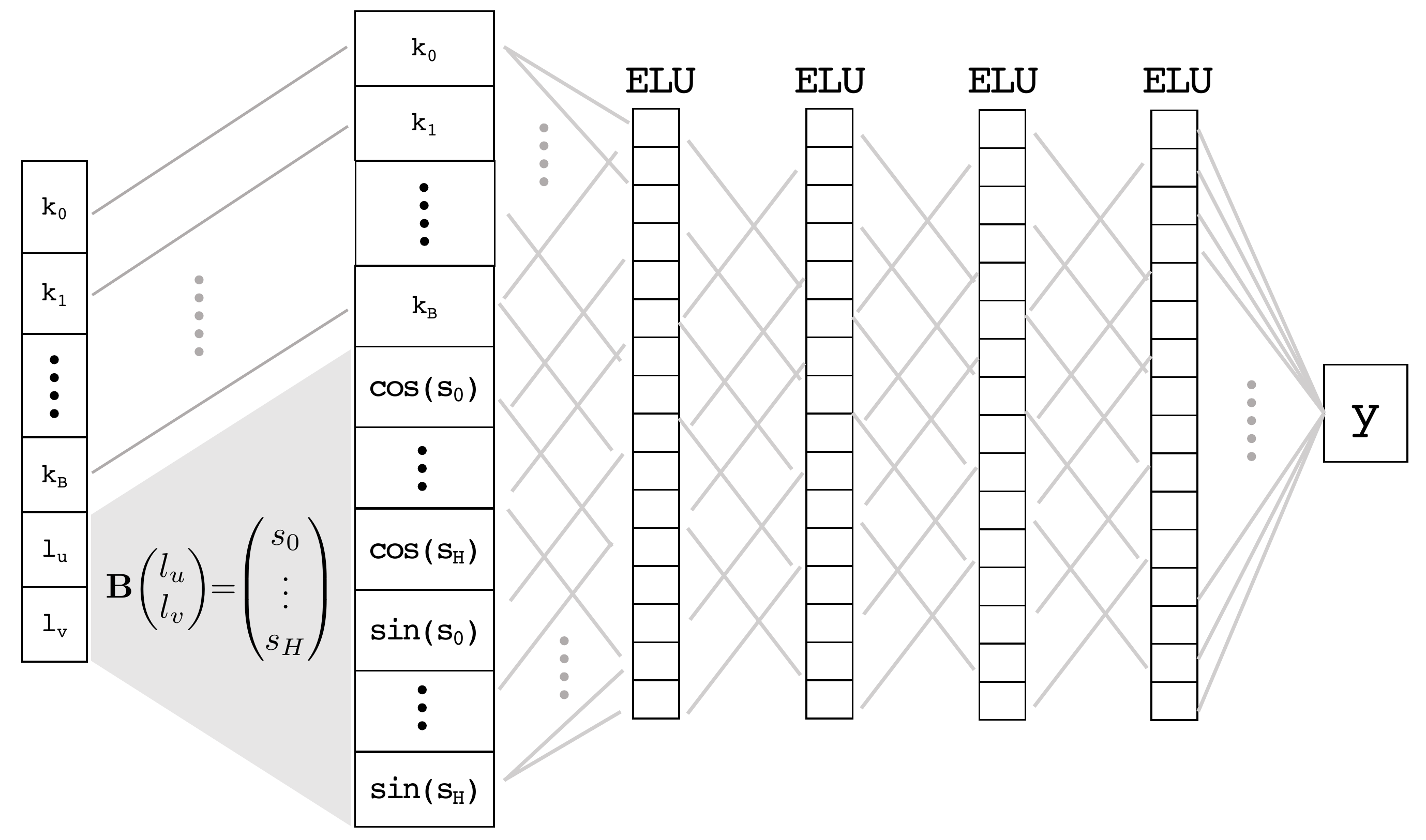}
    \caption{Left: 2D plot of the first two components of the light direction vectors $(\mathcal{L}_0 \ldots \mathcal{L}_n)$. Each point is associated to an image in the MLIC. Note the circular trajectory. Right: Network architecture composed by 5 fully-connected layers with ELU activation.}
    \label{fig:network_and_lights}
\end{figure}

Our proposed neural model $\mathcal{Z}(\mathbf{k}_p, \vec{l}) \to y$ works independently for each pixel (i.e. it does not consider the spatial relationship among those) and recovers the intensity information $y$ without the colour.
It takes as input a compressed light vector $\mathbf{k}_p = (k_0, k_1, \ldots, k_B) \in \mathbb{R}^B$ of any pixel $p$ and a light direction $\vec{l}$ to produce the intensity for pixel $p$. 

The model $\mathcal{Z}$ is composed by an initial (non-trainable) projection of the light vector followed by a MLP arranged in 5 layers consisting in $16$ neurons each, all using the ELU activation function except for the output realised with a single neuron with linear activation (Fig.~\ref{fig:network_and_lights}, right).
The network input $I$ is a $(B+2H)$-dimensional vector created by concatenating the $B$ values of $\mathbf{k}_p$ with the projection of $\vec{l}$ to an $H$-dimensional Fourier space with random frequencies.

Specifically, let $\mathbf{B}$ be a $H{\times}2$ matrix where each element is sampled from a Gaussian distribution $\mathcal{N}(0,\sigma^2)$.
This matrix is generated once for each acquired object, it is not trained, and is common to all the pixels. The network input is then obtained as:

\begin{equation}
    I = \begin{pmatrix} k_0, & \ldots & k_B, & \cos(s_0), & \ldots \cos(s_H), & \sin(s_0) & \ldots & \sin(s_H) \end{pmatrix}
\end{equation}

\begin{equation}
  \text{where} \hspace{.5cm} \begin{pmatrix}s_0 & \ldots s_H & \end{pmatrix}^T = \mathbf{B} \begin{pmatrix}l_u \\ l_v \end{pmatrix}.
\end{equation}

\noindent Once $\mathcal{Z}$ is trained, it can be used for relighting as follows:
\begin{equation}
    f\big( \mathbf{p}, \vec{l} ) = \bigg(\mathcal{Z}(\mathbf{k}_\mathbf{p}, \vec{l}),\;\bar{U}(\mathbf{p}),\; \bar{V}(\mathbf{p})\bigg).
\end{equation}

\subsubsection{Creating the compressed light vectors $k_p$}

The size of the neural model $\mathcal{Z}$ depends by the $2H$ values of the matrix $\mathbf{B}$, its internal weights, and the light vectors $\mathbf{k}_p$ (one for each pixel, for a total of $W{\cdot}H{\cdot}B$ values). It is obvious that most of the storage is spent for the light vectors since the number of image pixels is far greater than the other variables. Considering that we acquire roughly $1$ minute of video at $30$ FPS, our MLIC is composed by ${\approx}2000$ images cropped to a size of $400{\times}400$ pixels for a total of $320$ MB. So, using all the acquired data as-is to define $f$ jeopardises the idea of doing interactive relighting directly on a mobile app or in a web browser.
Since we assume that all the pixels observe the same light vector, the acquired MLIC $(\mathcal{I}_0 \ldots \mathcal{I}_N)$ can be represented as a $W{\times}H{\times}N$ $N$-channel image in which, for each pixel, a vector of $N$ values (corresponding to light directions $\mathcal{L}_0 \ldots \mathcal{L}_N$) have been observed.  In \cite{dulecha2020neural} the authors use an auto-encoder to produce an intermediate encoded representation of the observed light vectors of each pixel, and then just the decoder for relighting. This works well for the light dome in which $N$ is typically less than $50$. We tried their approach with $N=1500$ lights and realised that the network struggles to converge to an effective encoded representation.

We propose a more classical approach in which the encoding of the light vectors is not based on Deep Learning. Let $\mathbf{K}_p$ be the $N$-dimensional light vector of the pixel $p$. We propose to use Principal Component Analysis on all the light vectors acquired $\mathbf{K}_{p_0} \ldots \mathbf{K}_{p_{W{\times}H}}$ to find a lower-dimensional space of $B$ orthogonal bases. Then, the encoded $\mathbf{k}_p$ is obtained by projecting $\mathbf{K}_p$ into that space. In the experimental section we show how the number of bases $B$ can be very small compared to $N$ while still producing high-quality results. 

\subsubsection{Network training and implementation details}

The network model $\mathcal{Z}$ is trained by associating each input $I$ with the expected reflectance intensity. Specifically, we combined the encoded vector $\mathbf{k}_p$ of each pixel, with all the possible light directions $\mathcal{L}_0 \ldots \mathcal{L}_N$ to produce the input $I_{x,y,n}$  ( $0{\leq}y{<}H$, $0{\leq}x{<}W$, $0{\leq}n{<}N$).
The output associated to $I_{x,y,n}$ is simply the value of $\mathcal{I}_n(x,y)$, that is the intensity value observed for light $n$ at pixel $(x,y)$.
This results in a total of $W{\cdot}H{\cdot}N$ data samples to be used for training.
Note that, regardless the amount of pixels (i.e. the image resolution) and light directions involved (i.e. number of frames in the video), the network architecture remains unchanged. Therefore, it is easy to compute how much storage is needed for the model, depending only on the number of PCA bases $B$ and image resolution. 
Considering the acquired data size discussed before, and supposing to use $B=8$ bases and $H=10$ frequencies, we have to store $400{\times}400{\times}8$ compressed light vectors (${\approx}5$ MB with single precision), and $1252$ values for network weights and $\mathbf{B}$.

Finally, we adopted a classical Mean Absolute Error (MAE) loss function:

\begin{equation}
   \mbox{MAE} = \frac{1}{W{\cdot}H{\cdot}N}\sum_{x,y,n} |\mathcal{Z}(\mathbf{k}_{\mathbf{p}=(x,y)}, \vec{l}_n) - \mathcal{I}_n(x,y)|.
\end{equation}

\section{Experimental Results}

\begin{figure}[t]
    \centering
    \includegraphics[height=3.9cm]{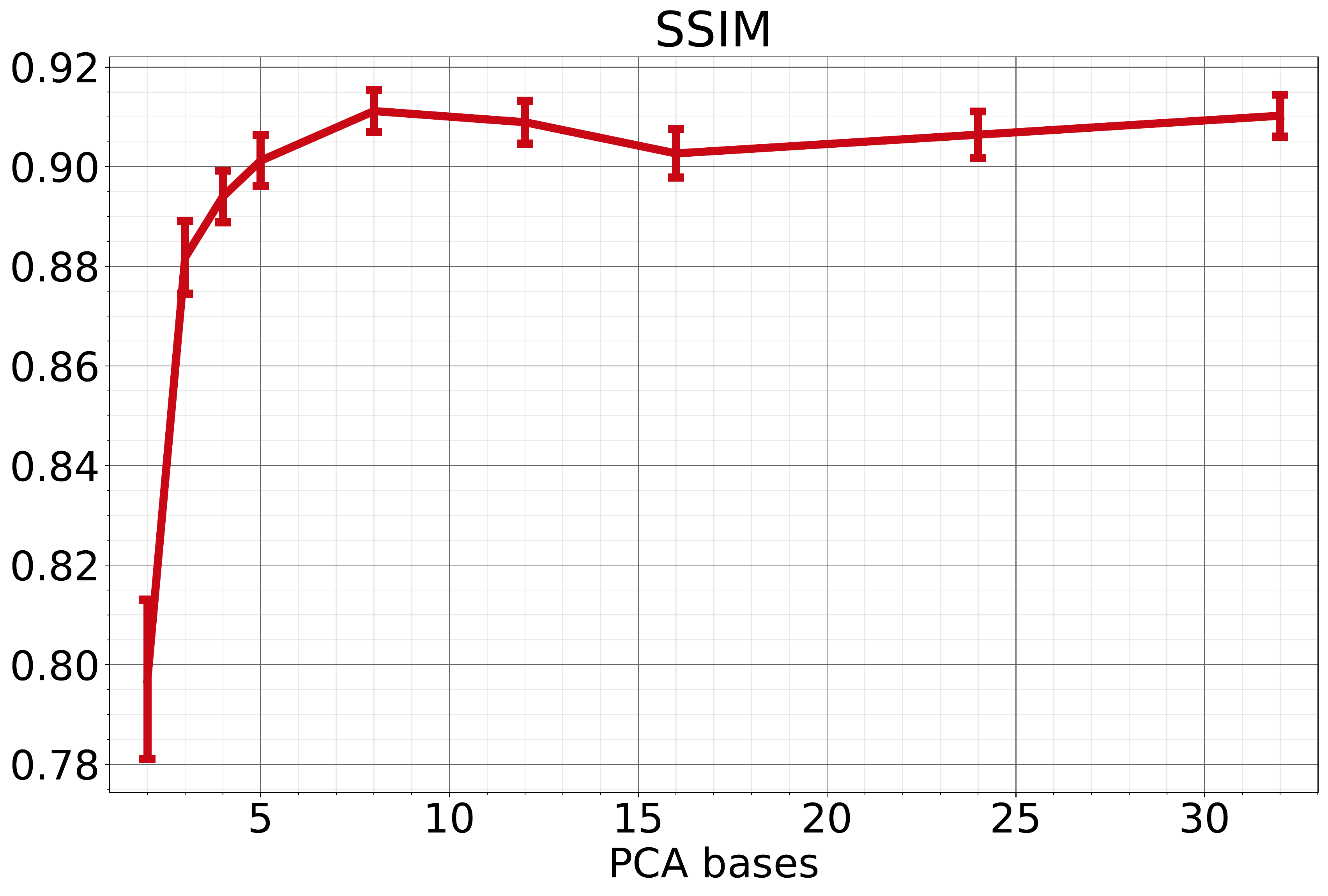}
    \includegraphics[height=3.9cm]{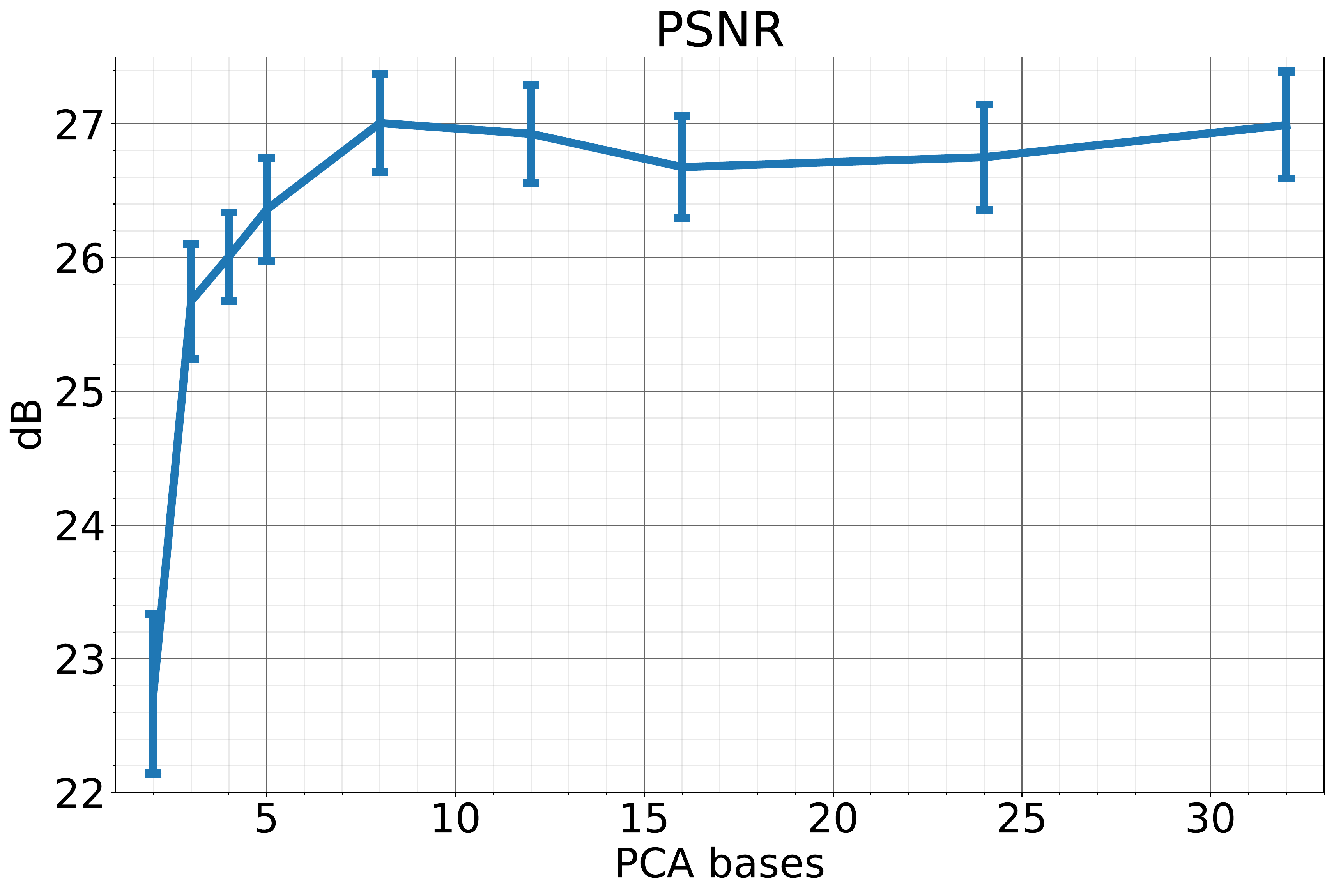}\\
    \includegraphics[height=3.9cm]{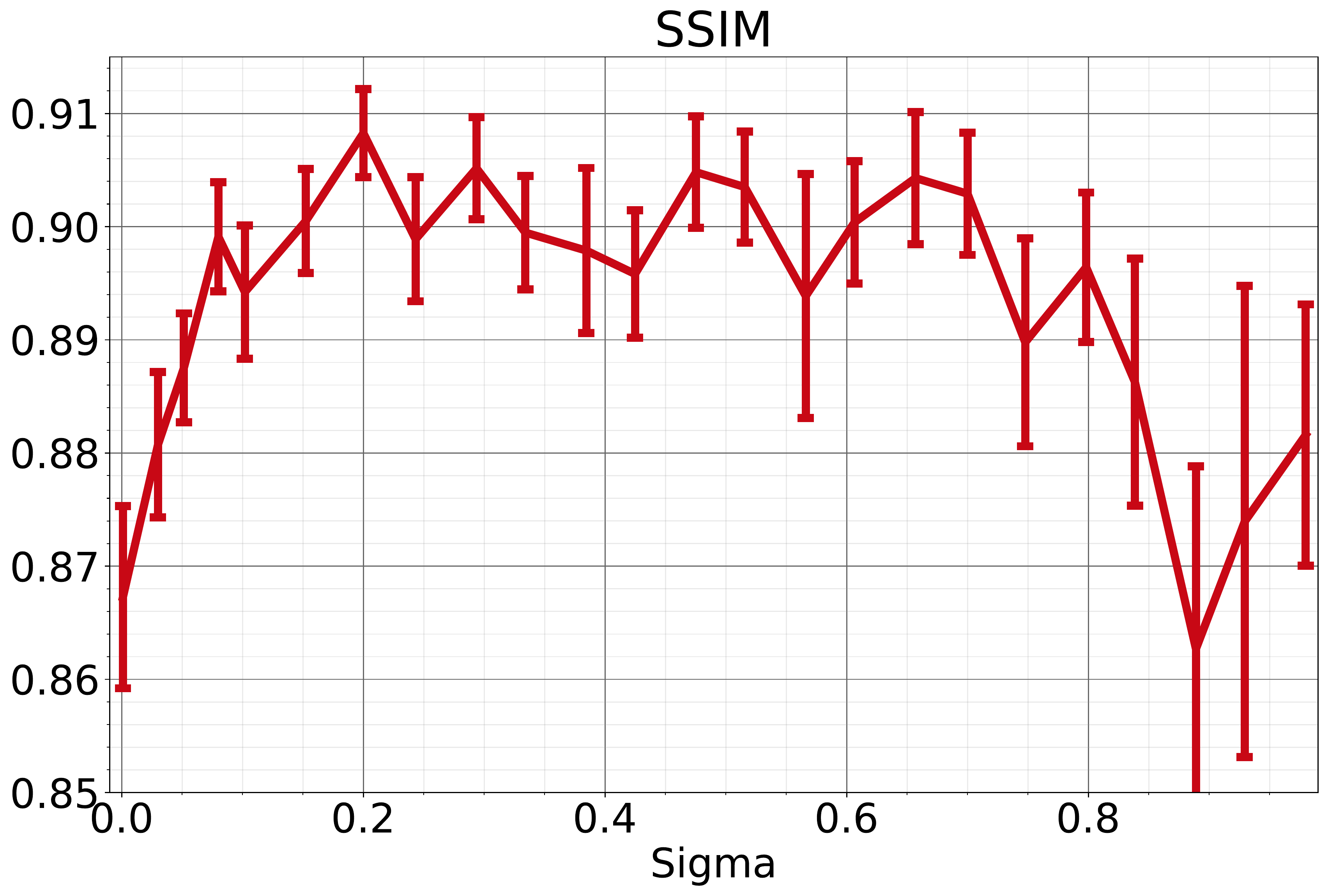}
    \includegraphics[height=3.9cm]{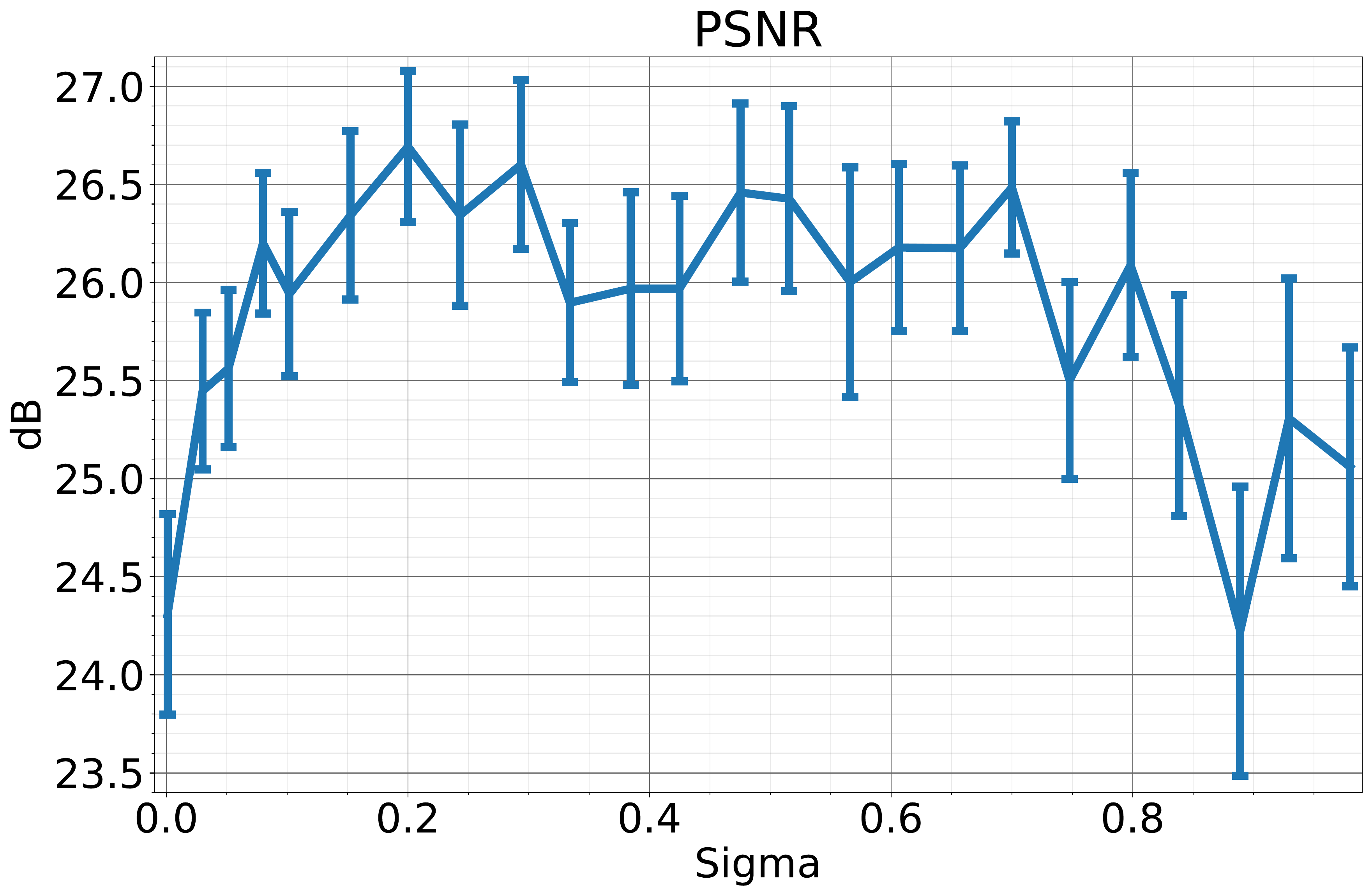}
    \caption{First row: SSIM and PSNR values increasing the number of PCA bases for data compression. Second row: SSIM and PSNR values increasing the sigma.}
    \label{fig:pca_sigma_exp}
\end{figure}

We started by analysing the behaviour of our proposed MLP model with respect to two relevant parameters, namely the number of PCA bases $B$ and the parameter $\sigma$ used to sample the frequencies of the light projection matrix $\mathbf{B}$.
Then, we quantitatively and qualitatively validated our method with respect to fully-synthetic data as well as with real-world smartphones acquisition.

In all our tests we fixed $H = 10$ so that the matrix $\mathbf{B}$ has size $10 \times 2$ always projecting  the input light vector $\vec{l}$ into a $20$-dimensional space.
Note that the values of the matrix $\mathbf{B}$ are randomly sampled before starting the training and never optimised.
During the training we used Adam optimiser with a learning rate of $10^{-3}$ for the first $20$ epochs and then reduced to $10^{-4}$ for another $20$.

\subsubsection{Real-world datasets}
For the real data we used some coins as test objects, acquired using an Apple iPhone 11 acting as static device and a Samsung Galaxy A40 as the moving one. Videos have been processed as described in Section \ref{sec:data_acquisition} resulting in a MLIC with roughly $2K$ images for each coin.
Then, for each MLIC we randomly selected $25$ lights for the test set and discarded the closest light directions (within a radius) so that the learning process is not trained on similar conditions.
An example of acquired light directions for the dataset \emph{Coin1} is shown in Figure \ref{fig:network_and_lights} (left), where the blue dots are the ${\sim}1920$ lights used for training and the red crosses the ones extracted for test.

\begin{figure}[t]
    \centering
    \includegraphics[height=3.9cm]{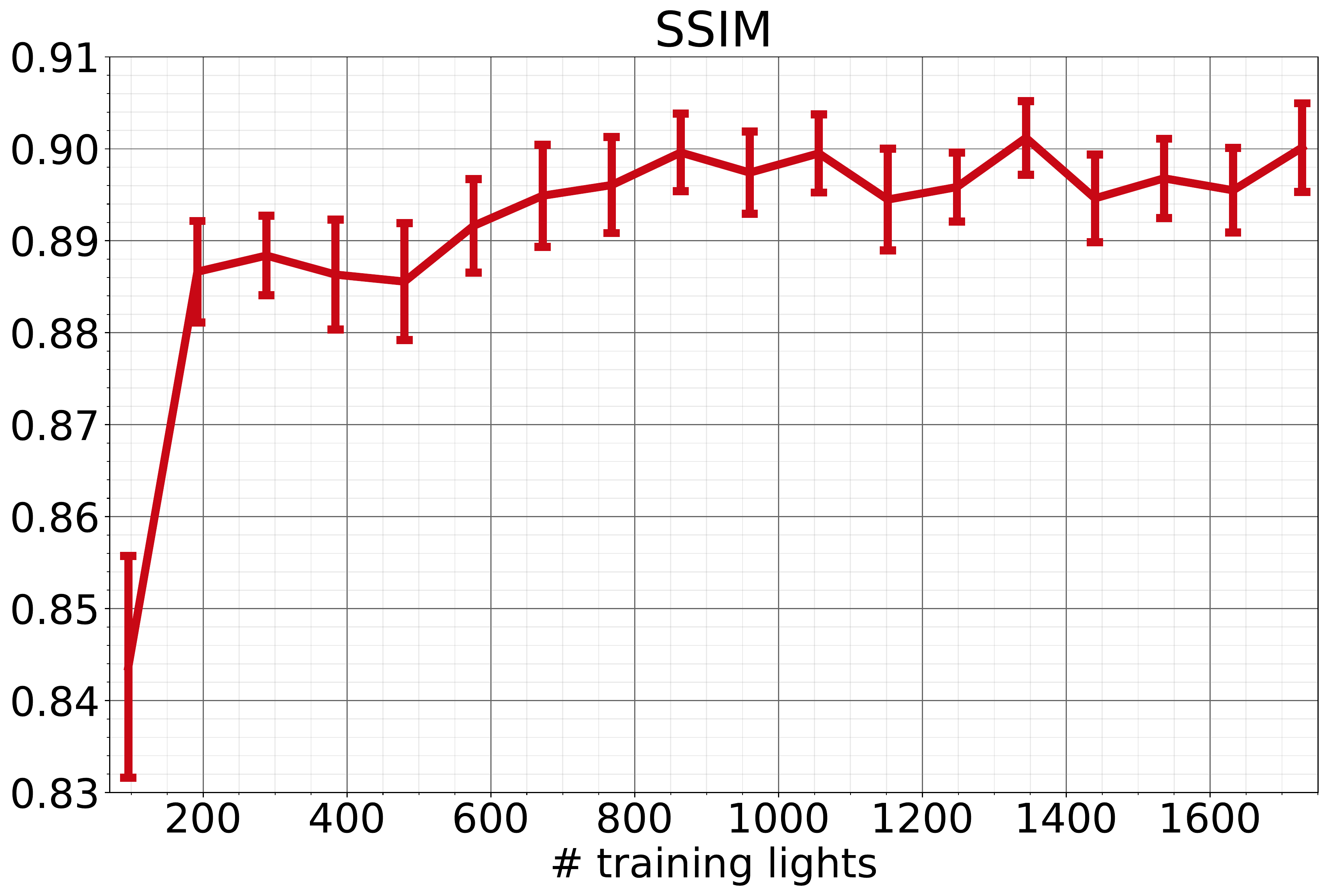}
    \includegraphics[height=3.9cm]{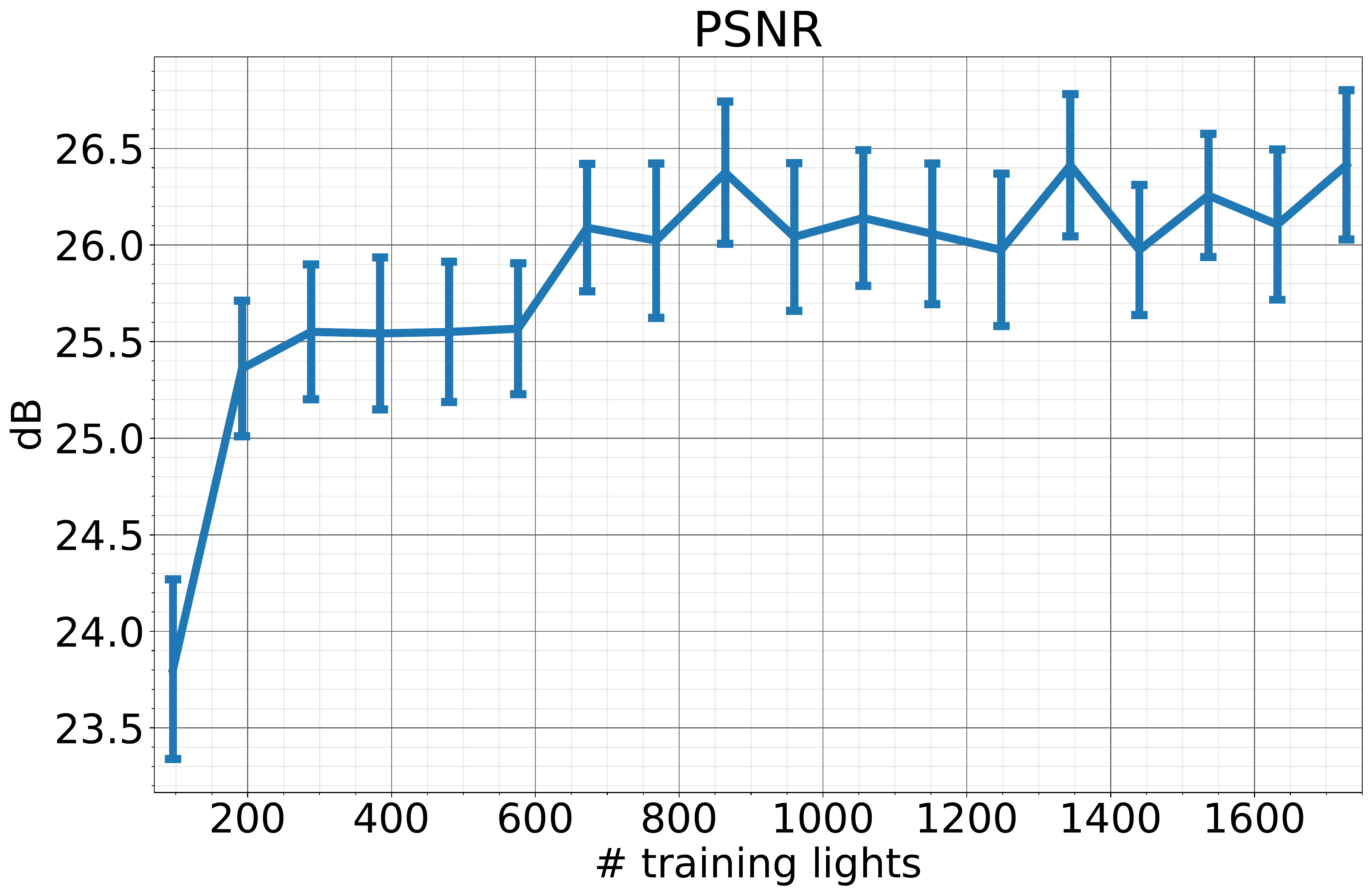}
    \caption{Average SSIM and PSNR on test set increasing the number of lights used during training procedure for our method.}
    \label{fig:n_lights_exp}
\end{figure}

\subsection{Parameter Study}

We first studied the effect of the number $B$ (i.e. PCA bases) on the final relighting quality.
Therefore, in the first test we projected the acquired MLIC data into an increasing number of PCA bases, and proceeded with the training process as described.
The plots in Figure \ref{fig:pca_sigma_exp} (first row) show the resulting PSNR (Peak Signal-to-Noise Ratio) and SSIM (Structural Similarity Index) for the test set while raising the number of bases $B$, from 2 to 32.
Note that such results have been computed by taking the average among all the acquired datasets and repeating the training 10 times due to the random nature of the process. The error bars denote the standard error.
We can notice that with $B = 2$ the relighting quality is quite low, and increasing the number of bases from 3 to 8 corresponds to an increasing reconstruction quality.
Both PSNR and SSIM value stabilise for $B > 8$, meaning that a higher number of bases would not further improve the output quality.
In all our tests we observed that a PCA projection with $8$ bases offers a good compression for our smartphone-acquired data.
Moreover, we tested the same $8$-bases compression for classical RTI datasets where a dome with equispaced lights is used: interestingly, results are numerically and qualitatively better with respect to the autoencoder compression technique as shown in the first row of Table~\ref{table:comparison}.

The next experiment results are shown in the second row of Figure \ref{fig:pca_sigma_exp}: we analysed the relighting quality against the value of $\sigma$ (on x-axis) used to generate the random values in the matrix $\mathbf{B}$.
The test was repeated 50 times for each different dataset.
We can observe that values around $\sigma = 0.3$ offer good results in terms of average PSNR and SSIM on the test set, exhibiting also a smaller standard error.
As stated in\cite{tancik2020fourier}, $\sigma$ is a free parameter that has to be tuned for a particular problem. However, our light directions have unitary norm so, once the optimal $\sigma$ is defined, it will remain the same regardless the object to reconstruct. Therefore, we used $0.3$ in all our real-world tests.

We also tested the effect of the number of acquired light directions (the size of $N$) against the final reconstruction accuracy while keeping the same network layout (Fig.\ref{fig:n_lights_exp}). This increases the size of the training set but not the storage space required for the model. Both SSIM and PSNR increase with $N$, probably because the network can be trained better if a large variety of light conditions can be used. Nevertheless, this increase is almost negligible when the number of light samples exceeds $700-800$. So, assuming an acquisition in which a carefully planned circular motion around the object is performed, an average video duration of $40$ seconds at $25$ FPS would be sufficient.

\subsection{Comparisons}

\setlength{\tabcolsep}{4pt}
\begin{table}[t]
\begin{center}
\caption{Relight comparison for different methods.}
\label{table:comparison}
\begin{tabular}{|c|c|c|c|c|c|c|c|c|} 
    \cline{2-9}
    \multicolumn{1}{c|}{} & \multicolumn{2}{c|}{\textbf{Polynomial}} & \multicolumn{2}{c|}{\textbf{RBF}} & \multicolumn{2}{c|}{\textbf{NeuralRTI}} & \multicolumn{2}{c|}{\textbf{Our}} \\
    \hline
    \textbf{Dataset} & PSNR & SSIM & PSNR & SSIM & PSNR & SSIM & PSNR & SSIM\\ 
    \hline \hline
    SynthRTI & $22.7451$ & $0.7932$ & $22.6828$ & $0.8353$ & $26.3658$ & $0.8540$ & $\mathbf{26.4075}$ & $\mathbf{0.8553}$\\ 
    \hline \hline
    Coin1 & 24.0562 & 0.8643 & 25.6791 & \textbf{0.9152} & 25.6846 & 0.8940 & \textbf{27.0019} & 0.9118\\ 
    \hline
    Coin2 & 25.8627 & 0.8798 & 26.8939 & \textbf{0.9197} & 26.9147 & 0.8937 & \textbf{28.0361} & 0.9105\\ 
    \hline
    Coin3 & 24.2319 & 0.8642 & 25.4360 & \textbf{0.9009} & 25.0304 & 0.8808 & \textbf{25.7479} & 0.8975\\ 
    \hline
    Coin4 & 27.6388 & 0.9155 & 28.1845 & 0.9369 & 27.8950 & 0.9176 & \textbf{29.6954} & \textbf{0.9398}\\ 
    \hline
\end{tabular}
\end{center}
\end{table}
\setlength{\tabcolsep}{1.4pt}

\begin{figure}
    \centering
    \begin{tabular}{ccccc}
    Polynomial & RBF & NeuralRTI & Our & GT\\
    \includegraphics[width=0.19\linewidth]{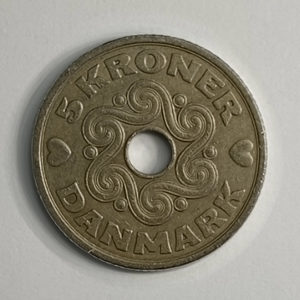}&
    \includegraphics[width=0.19\linewidth]{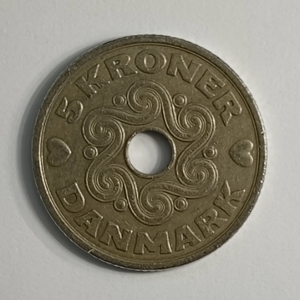}&
    \includegraphics[width=0.19\linewidth]{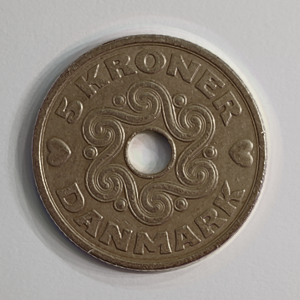}&
    \includegraphics[width=0.19\linewidth]{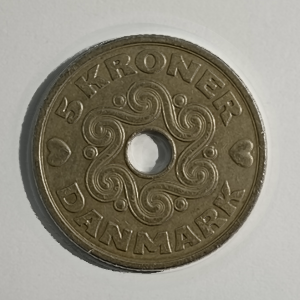}&
    \includegraphics[width=0.19\linewidth]{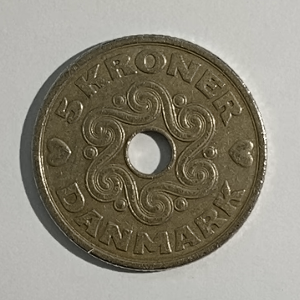}\\
    \includegraphics[width=0.19\linewidth]{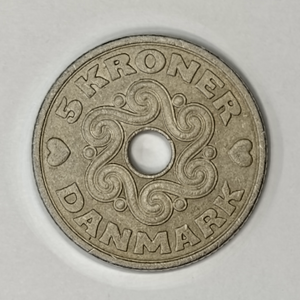}&
    \includegraphics[width=0.19\linewidth]{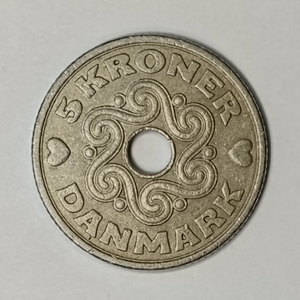}&
    \includegraphics[width=0.19\linewidth]{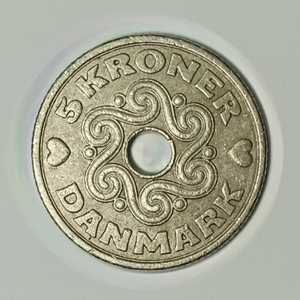}&
    \includegraphics[width=0.19\linewidth]{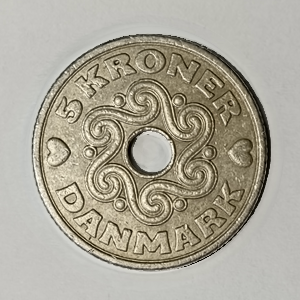}&
    \includegraphics[width=0.19\linewidth]{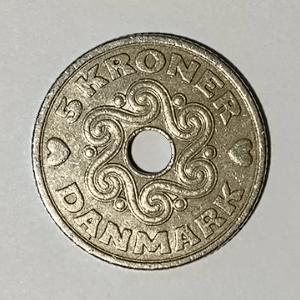}\\
    \includegraphics[width=0.19\linewidth]{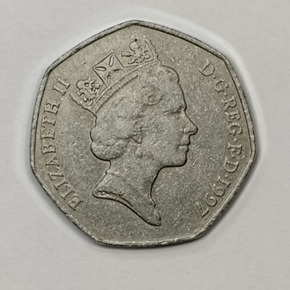}&
    \includegraphics[width=0.19\linewidth]{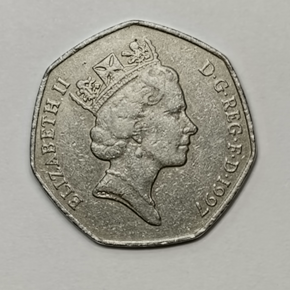}&
    \includegraphics[width=0.19\linewidth]{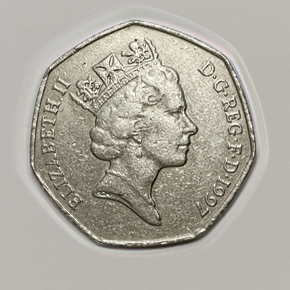}&
    \includegraphics[width=0.19\linewidth]{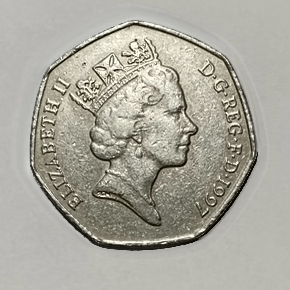}&
    \includegraphics[width=0.19\linewidth]{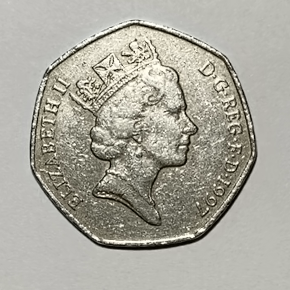}\\
    \includegraphics[width=0.19\linewidth]{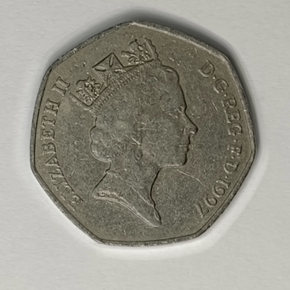}&
    \includegraphics[width=0.19\linewidth]{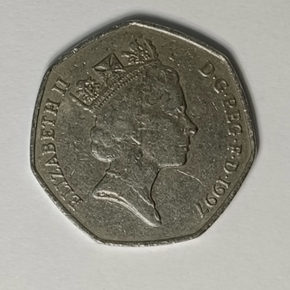}&
    \includegraphics[width=0.19\linewidth]{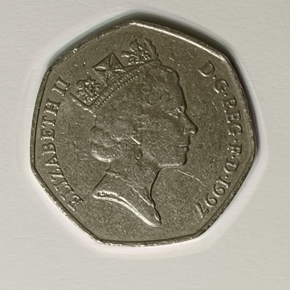}&
    \includegraphics[width=0.19\linewidth]{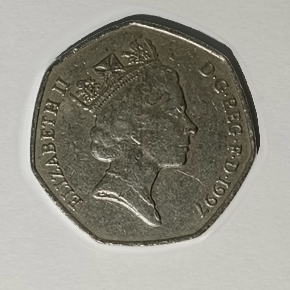}&
    \includegraphics[width=0.19\linewidth]{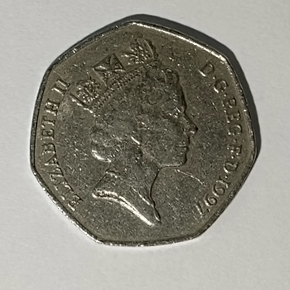}\\
    \includegraphics[width=0.19\linewidth]{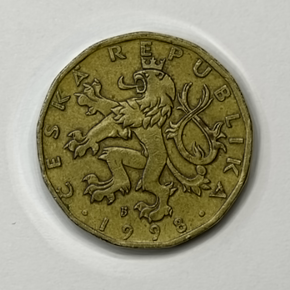}&
    \includegraphics[width=0.19\linewidth]{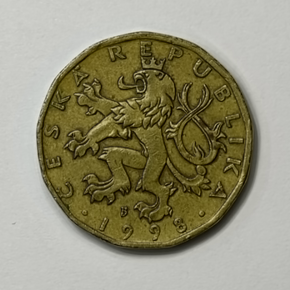}&
    \includegraphics[width=0.19\linewidth]{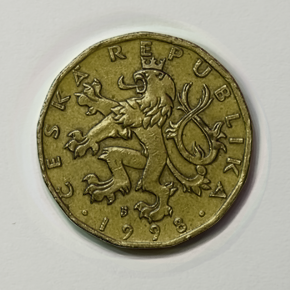}&
    \includegraphics[width=0.19\linewidth]{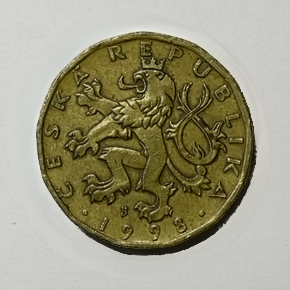}&
    \includegraphics[width=0.19\linewidth]{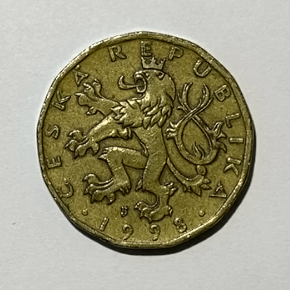}\\
    \includegraphics[width=0.19\linewidth]{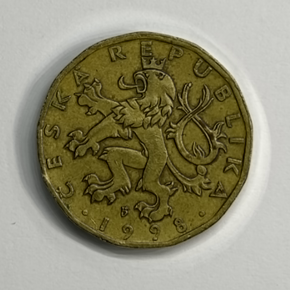}&
    \includegraphics[width=0.19\linewidth]{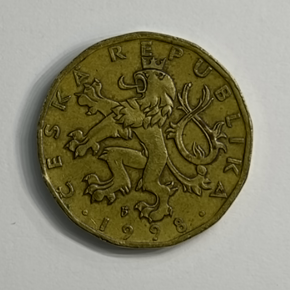}&
    \includegraphics[width=0.19\linewidth]{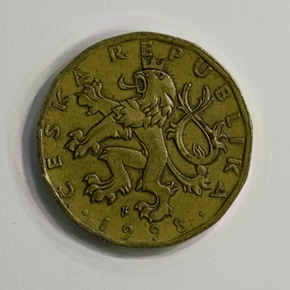}&
    \includegraphics[width=0.19\linewidth]{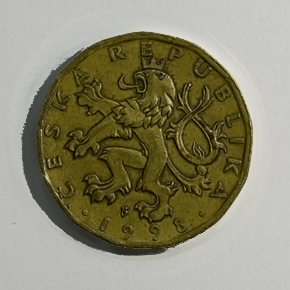}&
    \includegraphics[width=0.19\linewidth]{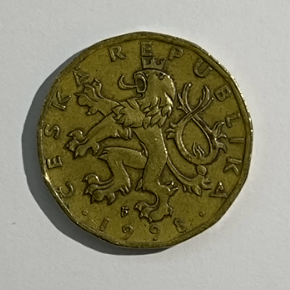}
    \end{tabular}
    \caption{Relighting comparison of real-world data acquired with two smartphones. The last column (ground truth, GT) shows the actual pictures from the test set.}
    \label{fig:real_comparison}
\end{figure}

We compared our relighting approach with two classical light interpolation methods, namely polynomial texture maps \cite{malzbender2001polynomial} (from now on, identified as \emph{polynomial}) and RBF.
Moreover, we tested against the already discussed learning-based method \emph{NeuralRTI} \cite{dulecha2020neural}.
We recall that NeuralRTI architecture changes with the number of input lights (the length of the encoder input is $3N$ because the network works in the RGB space), so we trained it on our acquired data by randomly selecting $100$ lights among the training set, since the training process becomes unfeasible for highest input dimensions.

In addition to our data acquired with smartphones, we also validated the method on classical RTI configurations represented by the synthetic dataset presented in \cite{dulecha2020neural}.
Such data is generated simulating a dome with $69$ lights, divided into separate train and test sets of $49$ and $20$ lights respectively.
In all comparisons we set $B=8$ (PCA bases), $\sigma = 0.3$ and $H = 10$.
Table \ref{table:comparison} shows the comparison results.
The values in the first row represent the average SSIM and PSNR for the whole SynthRTI dataset. To better evaluate our method comprising not only the reflectance model but also the smartphone-based data acquisition, we show the results for all the objects of the real-world dataset acquired as proposed.
Overall, our method exhibits the higher PSNR value, while in some cases relighted data interpolated with RBF give a slightly higher SSIM, but with a significantly smaller PSNR with respect to our method. Note however that RBF is significantly slower in the relighting phase.
Also, our values are slightly better with respect to NeuralRTI for the synthetic dataset, where the training lights are sampled uniformly on a dome setup. This indicates that our proposed PCA compression and decoder network still improves the encoder-decoder architecture of~\cite{dulecha2020neural}.
Note that we did not tune any parameter for our results, concluding that the number of PCA bases does not depend on the specific dataset.

\begin{figure}[t]
    \centering
    \begin{tabular}{ccccc}
    Polynomial & RBF & NeuralRTI & Our & GT\\
    \includegraphics[width=0.19\linewidth]{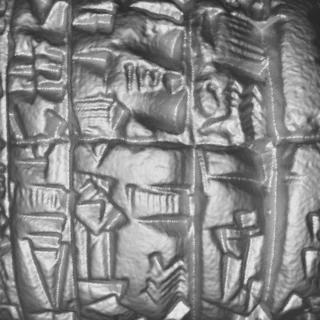}&
    \includegraphics[width=0.19\linewidth]{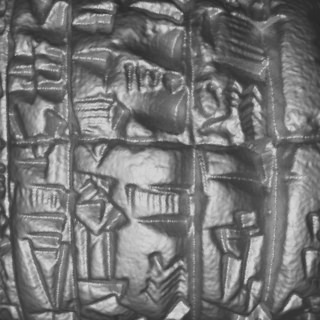}&
    \includegraphics[width=0.19\linewidth]{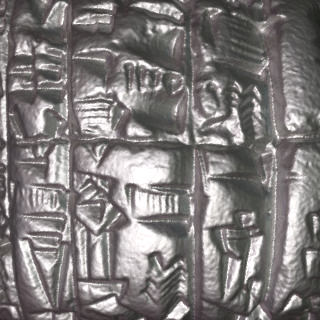}&
    \includegraphics[width=0.19\linewidth]{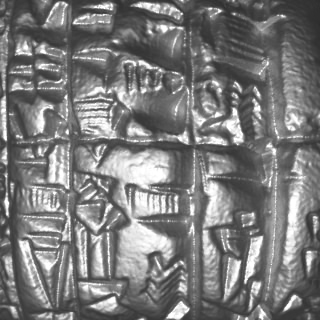}&
    \includegraphics[width=0.19\linewidth]{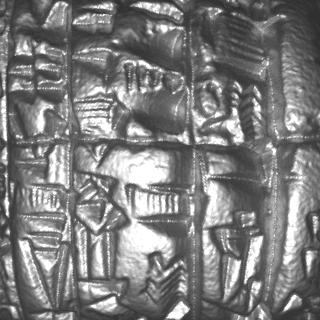}\\
    \includegraphics[width=0.19\linewidth]{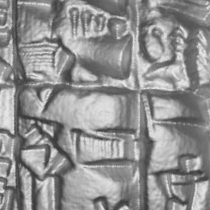}&
    \includegraphics[width=0.19\linewidth]{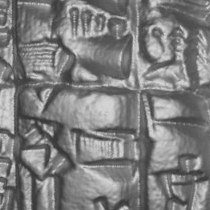}&
    \includegraphics[width=0.19\linewidth]{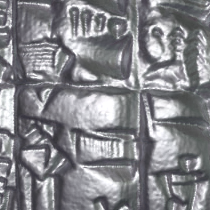}&
    \includegraphics[width=0.19\linewidth]{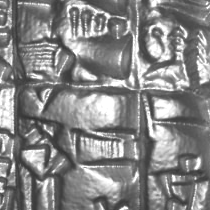}&
    \includegraphics[width=0.19\linewidth]{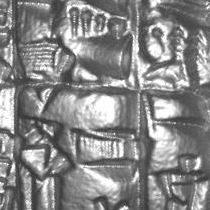}\\
    \includegraphics[width=0.19\linewidth]{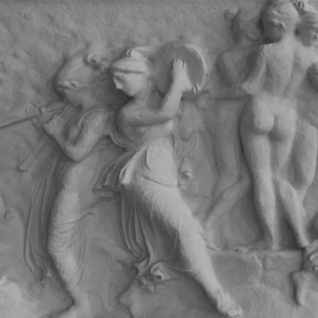}&
    \includegraphics[width=0.19\linewidth]{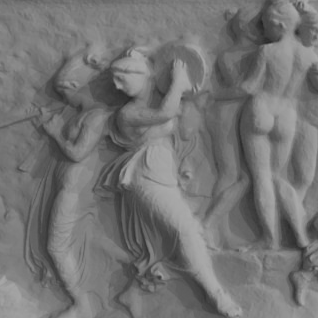}&
    \includegraphics[width=0.19\linewidth]{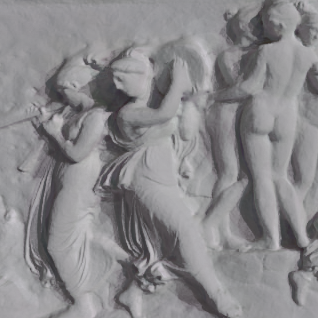}&
    \includegraphics[width=0.19\linewidth]{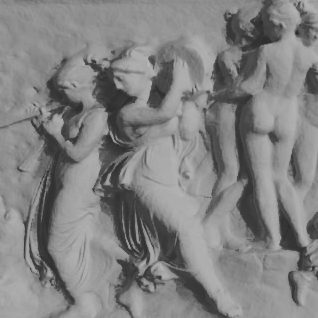}&
    \includegraphics[width=0.19\linewidth]{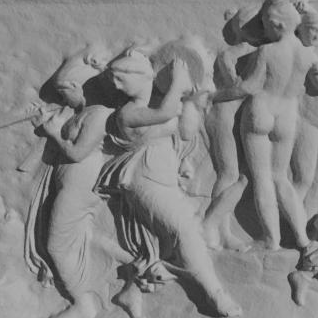}\\
    \includegraphics[width=0.19\linewidth]{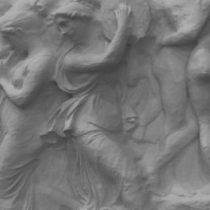}&
    \includegraphics[width=0.19\linewidth]{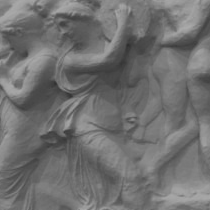}&
    \includegraphics[width=0.19\linewidth]{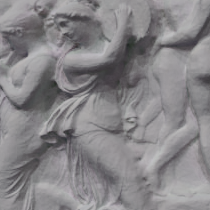}&
    \includegraphics[width=0.19\linewidth]{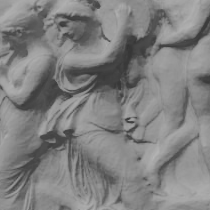}&
    \includegraphics[width=0.19\linewidth]{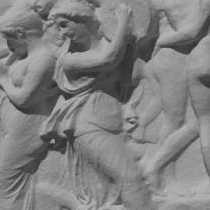}\\
    \end{tabular}
    \caption{Qualitative comparison on synthetic data generated with a dome configuration.}
    \label{fig:synth_comparison}
\end{figure}

Qualitative examples for our acquired dataset are shown in Figure~\ref{fig:real_comparison}, where we display the relighting of three coins with two different test lights (last column shows the ground truth, GT).
We can notice that our method is able to recover the object reflectance with high accuracy, especially for the shadows projected near the coins, while the other methods tend to generate light blooms or blurry shadows.
Moreover, we notice that NeuralRTI slightly alters the output tint with respect to the original: this can be seen in particular in the first two rows. Probably, directly modelling each pixel intensity and colour is more difficult to handle for the network than just the intensity. Using the average UV-value is easier and produces more stable results for non-iridescent objects.
Finally, in Figure~\ref{fig:synth_comparison} we show a couple of outputs for the synthetic dataset.
Our results are quite similar with respect to NeuralRTI but also in this case our shadow areas are sharper and the images exhibit a higher contrast.

\section{Conclusions}

In this paper we proposed a low-cost technique to perform image relighting on the go using two smartphones for data acquisition.
A practical video processing pipeline extracts the MLIC that is compressed and used to train a neural relighting model.
Extensive tests in both synthetic and real-world settings show that our network effectively hallucinates images from unseen light directions with high quality.
The presented setup can be easily operated directly on the field, with no need of expensive and specialised hardware, allowing researchers to carry out part of their work in an effective and fast way.

%
%
\bibliographystyle{splncs04}
\bibliography{egbib}
\end{document}